%% file: paper.tex
\documentclass[]{amap}
\usepackage{amssymb}

\usepackage[toc,page,header]{appendix}
\usepackage{minitoc}
\usepackage{solarized-light}
\usepackage{booktabs}
\usepackage{multirow}
\usepackage{graphicx}
\usepackage{array}
\usepackage{siunitx,array}
\usepackage[table]{xcolor}
\usepackage{makecell,array,booktabs}
\usepackage{makecell}
\usepackage{framed}

\usepackage{bbding}
\usepackage{hyperref}
\usepackage{amsmath}
\usepackage{amsfonts}
\usepackage{placeins}
\usepackage[colorinlistoftodos]{todonotes}
\usepackage{longtable}
\usepackage{hhline}
\usepackage{fancyvrb}
\usepackage{graphicx}
\usepackage[table]{xcolor}
\usepackage{booktabs}
\usepackage{float}
\usepackage{fvextra}
\usepackage{CJKutf8}
\usepackage{multicol}
\usepackage{float}
\usepackage{placeins}
\usepackage{cleveref}
\usepackage{tablefootnote}
\usepackage{threeparttable}
\usepackage{tabularx}
\usepackage{mdframed}
\usepackage{subcaption}
\usepackage[usestackEOL]{stackengine}
\usepackage[numbers]{natbib}
\newcommand{\commentout}[1]{}
\renewcommand{\paragraph}[1]{\noindent\textbf{#1.}\hspace*{1em}}
\usepackage{enumitem}
\usepackage{hyperref}
\setlist[itemize]{leftmargin=15pt}
\usepackage{siunitx,array}
\usepackage{tabularx}
\usepackage{adjustbox}
\usepackage{arydshln}
\usepackage{pifont}
\usepackage{bbding}
\definecolor{ampblue}{rgb}{0.82, 0.88, 0.94}
\usepackage[table]{xcolor} % 用于颜色
\usepackage{array} 
\usepackage[dvipsnames]{xcolor}

\RequirePackage{xspace}
\makeatletter
\DeclareRobustCommand\onedot{\futurelet\@let@token\@onedot}
\def\@onedot{\ifx\@let@token.\else.\null\fi\xspace}

\makeatother

\newcommand{\method}{\textit{ABot-Earth 0.5}\xspace}
\newcommand{\abotgs}{ABot-3DGS\xspace}

\crefname{figure}{Fig.}{Figs.}  % 小写形式（单数，复数）
\Crefname{figure}{Fig.}{Figs.}  % 首字母大写形式（单数，复数）

\crefname{table}{Tab.}{Tabs.}
\Crefname{table}{Tab.}{Tabs.}
\Crefname{section}{Sec.}{Secs.}
\Crefname{section}{Sec.}{Secs.}

\usepackage{xcolor}

\definecolor{abot1}{HTML}{0185FE}
\definecolor{abot2}{HTML}{0185FE}
\definecolor{abot3}{HTML}{0185FE}
\definecolor{abot4}{HTML}{0185FE}
\definecolor{abot5}{HTML}{FB8C00}
\definecolor{abot6}{HTML}{FB8C00}
\definecolor{abot7}{HTML}{FB8C00}

\title{ABot-Earth 0.5: Generative 3D Earth Model}

\author{AMAP CV Lab}
\vspace{-40pt}

\abstract{
We present \method{}, a generative 3D framework designed to synthesize vast, seamless 3D environments from ubiquitous, geospatially referenced satellite imagery. To achieve this, we propose a novel generative model formulated directly with the 3D Gaussian Splatting (3DGS) representation. The model is trained on a diverse corpus of existing real-world urban reconstructions, learning to generate realistic geometry and textures. At inference, it synthesizes novel 3D scenes conditioned solely on satellite imagery at a scalable rate of under 10 minutes per square kilometer, while demonstrating exceptional realism. The framework is designed for accessibility, with integrated hierarchical level-of-detail (LOD) structures that permit real-time, interactive visualization on web-based map engines. This high-fidelity simulation sandbox effectively mitigates the sim-to-real domain gap, enabling critical downstream Embodied AI applications like closed-loop UAV navigation. By providing an ultra-low-cost and high-efficiency solution, \method{} significantly lowers the technical and financial barriers to large-scale 3D reconstruction and empowers the future of global digital earth visualization.

\bigskip

\textbf{Official Page: } \url{http://abot-earth.amap.com/}

}

\begin{document}
\maketitle
\vspace{-4pt}

\begin{figure}[H]
\centering
\includegraphics[width=\textwidth]{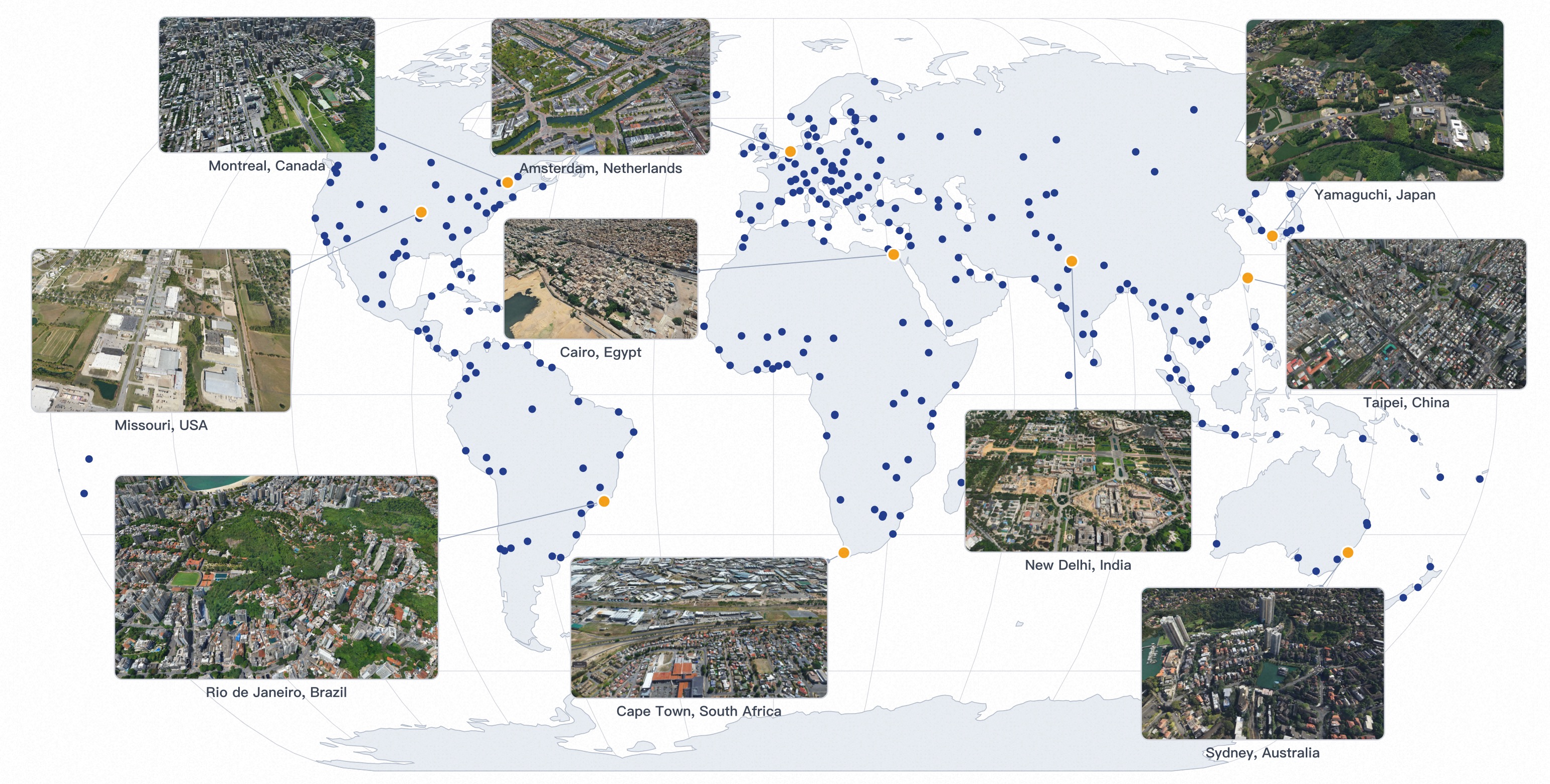}
\caption{We are unveiling \method{}, a generative 3D Earth model. Our official launch showcases an evolving 3DGS world spanning over 300 cities across 190+ countries, with continuous global expansion.}

\label{fig:teaser}
\end{figure}

\newpage
% \tableofcontents
% \newpage

%\newpage
\input{sections/sec1_introduction}
\input{sections/sec2_data_pipeline}
\input{sections/sec3_method}
\input{sections/sec4_deployment}
\input{sections/sec5_evaluation}
\input{sections/sec6_conclusion}

% \clearpage
\input{sections/contribution}

% \clearpage

% \bibliographystyle{plainnat}
% % \bibliography{main}
% \bibliography{reference}

% \bibliographystyle{unsrtnat}
% % \bibliography{main}
% \bibliography{reference}

\bibliographystyle{IEEEtran}
% \bibliography{main}
\bibliography{reference}

% \newpage
% \input{sections/sec7_appendix}
% 
% \clearpage
% \clearpage
% % \beginappendix

% \input{sections/appendix}
% \let\clearpage\relax

\end{document}

%% file: sections/sec1_introduction.tex
\section{Introduction}
\label{sec:intro}
% (Align to https://arxiv.org/pdf/2601.20540)

High-fidelity three-dimensional geospatial reconstruction of the Earth's surface has emerged as a foundational pillar for modern digital twin infrastructures, smart city logistics, and virtual simulation ecosystems. In particular, aerial-view 3D representations provide critical geospatial priors for rapid disaster response, urban planning, and robotic exploration. Despite its immense utility, traditional large-scale 3D reconstruction pipelines~\cite{schoenberger2016sfm,schoenberger2016mvs,nex2014uav,WEHR199968,shan2018topographic}, primarily built on dense oblique photogrammetry and LiDAR scanning, are fundamentally constrained by extreme data acquisition costs, prolonged processing latencies, and high computational barriers, rendering real-time or on-demand planetary-scale modeling a long-standing challenge.

As a compelling alternative, generative 3D modeling offers a way to bypass these physical constraints by shifting the technical burden from exhaustive multi-view acquisition to learned structural priors. By leveraging implicit geometric knowledge, generative paradigms can synthesize complete 3D structures from highly sparse inputs, drastically reducing both data collection overhead and optimization latency. Under this paradigm, 3D generative modeling has reached remarkable maturity at the individual object scale, enabling the rapid, high-fidelity synthesis of diverse single assets~\cite{xiang2025native,xiang2025structured,CLAY,hunyuan3d22025tencent,lai2025hunyuan3d25highfidelity3d,Seed3D,gao2022get3d,jun2023shapegeneratingconditional3d}. Inspired by these object-level successes, the generative AI community has actively embarked on scaling these capabilities to the far more challenging domain of unbounded, large-scale outdoor scenes~\cite{earthcrafter,CityDreamer,sat2city,lin2023infinicity,sat2scene,Yang_2023_ICCV,Qian_2023_Sat2Density,sat2densitypp,qian2026sat3dgen,infinite_nature_2020}. 

However, directly migrating object-centric formulations~\cite{xiang2025native,xiang2025structured,CLAY,hunyuan3d22025tencent,lai2025hunyuan3d25highfidelity3d,Seed3D} to large-scale scene synthesis remains highly non-trivial. A major limitation of existing outdoor generators is their heavy reliance on synthetic virtual assets or unconstrained imaginary scene hallucination~\cite{Xie_2025_CVPR,CityDreamer,lin2023infinicity,sat2city,sat2scene,earthcrafter}. Because these generated environments are inherently artificial and lack real-world physical and geospatial authenticity, they fail to bridge the severe sim-to-real domain gap~\cite{8202133}, making them impractical for rigorous downstream simulation and real-world transfer. To address this issue, we advocate for generating native 3D scenes trained directly on high-quality real-world reconstructions. Among various 3D representations, 3DGS~\cite{3DGS} has revolutionized photometric reconstruction of real-world environments with its unmatched rendering fidelity, natively capturing complex non-manifold topologies like dense foliage, building facades, and specular water surfaces~\cite{liu2024citygaussian}. 

To scale this real-world-trained generative paradigm to unbounded, planetary-wide extents, establishing a scalable and globally consistent conditioning signal is paramount. In this context, globally ubiquitous and geospatially referenced remote sensing imagery serves as the ideal conditioning blueprint. By establishing a direct geospatial correspondence to the underlying terrain at virtually any coordinates on Earth, leveraging this pervasive modality unlocks the potential for infinite-scale, progressive 3D generation. In this work, we present \method{}, a generative 3D framework designed to realize this vision, enabling the synthesis of vast, near-seamless 3D aerial scenes conditioned on standard satellite imagery, without requiring knowledge of its precise acquisition angles or multi-view overlaps. By framing 3D generation directly in a native 3DGS generative space, \method{} preserves exceptional visual quality while achieving a scalable generation rate of under 10 minutes per square kilometer. Extensive quantitative and qualitative evaluations demonstrate that \method{} outperforms state-of-the-art baselines by a significant margin, exhibiting a clear superiority in rendering realism and geometric fidelity. To showcase its performance and versatility, the key capabilities of \method{} are highlighted as follows:

\begin{itemize}
    \item \textbf{Generation of Real-World Geospatial Complexity.} \method{} achieves remarkable improvements in capturing real-world complexity without relying on artificial heuristics or synthetic assets. Trained directly on diverse real-world urban reconstructions, the model is capable of robustly synthesizing highly detailed structures, such as intricate building facades, dense vegetation canopy, and coherent road networks, with natural textures. We successfully validate the model's performance across the vast majority of global metropolitan areas as well as numerous non-urban natural terrains, demonstrating its robust Earth-scale generalizability. Crucially, as a native 3DGS-based generative framework, our synthesized environments can be seamlessly integrated and co-edited with precisely reconstructed, high-fidelity 3DGS landmark models. This composability allows users to insert meticulously scanned real-world landmarks into generated contextual backdrops, yielding an exceptionally immersive, hybrid-reality experience.
    
    \item \textbf{Planetary-Scale Online Exploration via Native Multi-LOD.} To support seamless online exploration across vast geographical extents, \method{} natively generates hierarchical level-of-detail (multi-LOD) outputs. When paired with our customized, YunJing-based 3DGS visualization engine, this capability enables dynamic, viewport-dependent tile scheduling and streaming of trillion-scale Gaussian primitives. Users can seamlessly zoom in from a planetary overview to fine-grained, street-level details with fluid, interactive frame rates. This natively integrated multi-LOD pipeline avoids the need for expensive post-reconstruction downsampling, providing a smooth and continuous interactive experience for global-scale digital earth and geographic information systems.
    
    \item \textbf{Simulation-Ready Capabilities for Downstream Applications.} Grounded in physical and photometric realism, the 3D environments synthesized by \method{} are not merely static visual models, but interaction-ready virtual sandboxes. By generating authentic 3D geometry and multi-view consistent textures, the model establishes a high-fidelity closed-loop simulation and training platform for Embodied AI, particularly for unmanned aerial vehicle (UAV) navigation, obstacle avoidance, and control~\cite{AirSim,FlightGoggles}. Ultimately, by providing an ultra-low-cost, high-efficiency generation pipeline, \method{} lowers the technical and financial barriers to large-scale 3D reconstruction, empowering global researchers and enterprises to realize scalable, cost-effective geospatial applications in smart city planning, environmental monitoring, and rapid disaster response.
\end{itemize}

Through this unified generative framework, \method{} effectively mitigates the synthetic-to-real domain gap and delivers robust generalization across diverse environments worldwide. Ultimately, we hope this ultra-low-cost, highly efficient solution will break down technological and financial barriers, bridging the ``3D digital divide'' and fostering global geospatial technological equity, thereby empowering scalable 3D digital earth visualization and simulation on a global scale.

%% file: sections/sec2_data_pipeline.tex
\section{Data Pipeline}
\label{sec:data_pipeline}

The quality ceiling of a 3D generative model is fundamentally governed by its training data.
Unlike video-based world models that curate internet-sourced footage~\cite{Sora2024},
ABot-Earth directly uses city-scale 3DGS scenes as training
data, produced by our reconstruction engine \abotgs.
We design a robust data engine spanning four stages: large-scale imagery collection from diverse sources, 3DGS reconstruction via \abotgs, spatial partitioning and multi-view rendering to produce training tiles, and multi-granularity quality assessment and curation.

\begin{figure}[H]
\centering
\includegraphics[width=\textwidth]{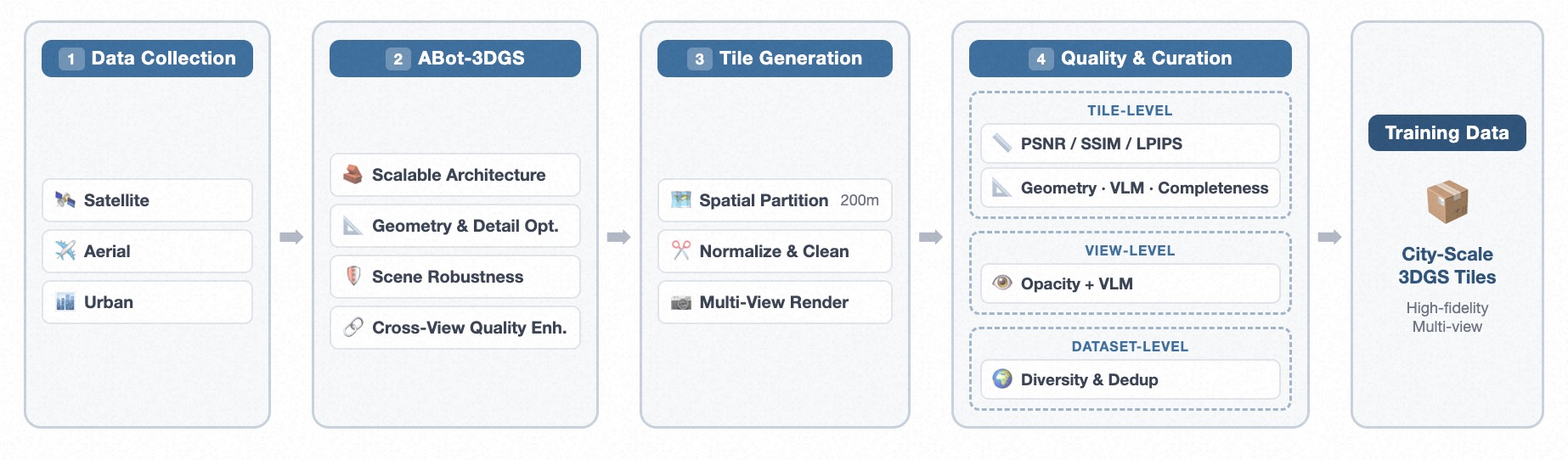}
\caption{Data pipeline overview. Multi-source imagery is
reconstructed into 3DGS scenes via ABot-3DGS, then spatially
partitioned into tiles and rendered from multi-view cameras. Multi-granularity quality
assessment and curation at the tile, view, and dataset levels ensure
only high-fidelity samples enter the final training set.}
\label{fig:data_pipeline}
\end{figure}

The overall pipeline is illustrated in \cref{fig:data_pipeline}.

\subsection{Data Collection}
\label{sec:data_acquisition}

\begin{figure}[t]
\centering
\includegraphics[width=\textwidth]{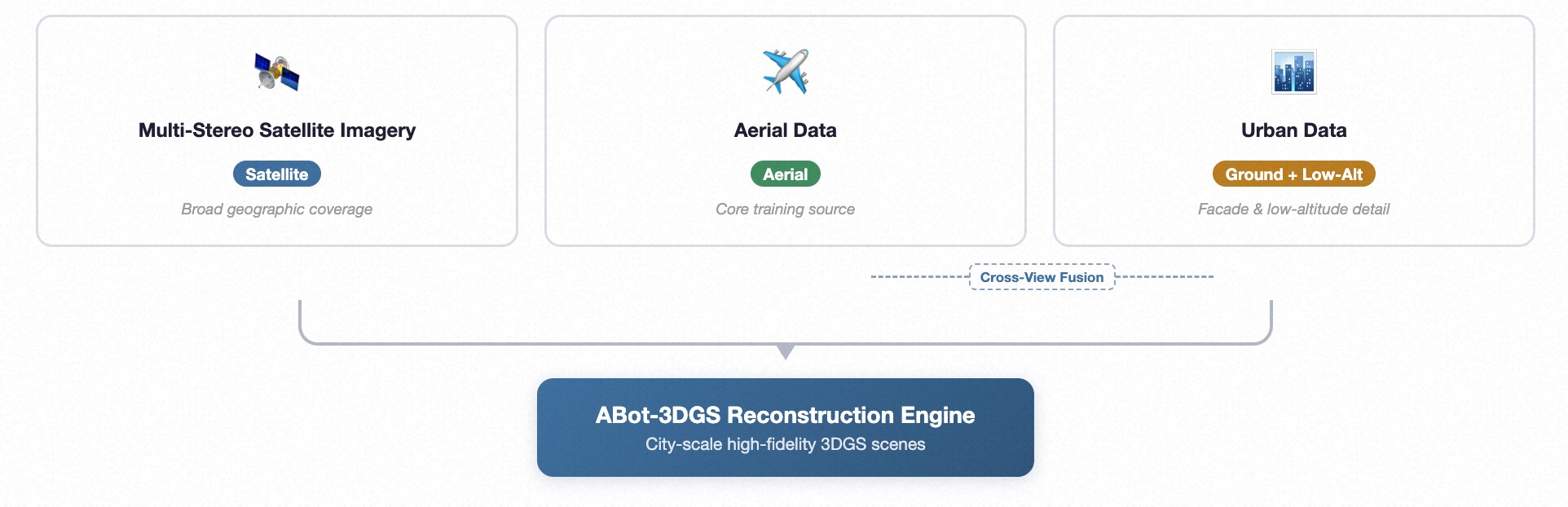}
\caption{Data sources overview. Satellite, aerial, and urban imagery
provide complementary viewpoint coverage for large-area reconstruction.
Each category combines proprietary and public sources.}
\label{fig:data_sources}
\end{figure}

To build the training foundation for 3DGS generation, we collect real-world imagery from three complementary categories: satellite, aerial, and urban (\cref{fig:data_sources}). Together they greatly expand data scale and scene diversity by spanning the full range from orbital to ground-level viewpoints, and further improve reconstruction quality in regions where cross-viewpoint coverage is available. Each category draws on both proprietary acquisitions and curated public datasets (Table~\ref{tab:open_datasets}); all sources are real-world captures rather than synthetic assets, and undergo unified coordinate transformation and metadata standardization before entering the \abotgs pipeline.

\paragraph{Multi-Stereo Satellite Imagery}
Multi-stereo satellite imagery consists of multiple orbital captures at varying off-nadir angles over the same region, providing the parallax necessary for 3D reconstruction. We collect such imagery from both proprietary acquisitions and public benchmarks (e.g., DFC 2019~\cite{DFC2019}; see Table~\ref{tab:open_datasets}) and reconstruct it into 3DGS scenes via FromOrbit2Ground~\cite{orbit2ground}, a satellite reconstruction module within ABot-3DGS.
FromOrbit2Ground addresses the extreme viewpoint gap between orbital captures and ground-level rendering through a two-stage pipeline: a Z-Monotonic SDF recovers watertight urban geometry from sparse top-down views, and a diffusion-based restoration network synthesizes high-fidelity facade textures.
This satellite-to-3DGS path significantly expands geographic coverage, allowing scalable acquisition of training scenes across diverse urban and natural landscapes worldwide.

\paragraph{Aerial Data}
We leverage high-resolution oblique aerial data as the core training source, covering built-up areas and natural landscapes across multiple cities and regions. After standardized preprocessing, the imagery feeds directly into the ABot-3DGS reconstruction pipeline, producing photorealistic 3DGS scenes. The pipeline optionally integrates LiDAR point clouds and pre-built photogrammetric meshes as auxiliary geometric priors to further improve surface reconstruction quality. We also incorporate public UAV datasets such as UrbanScene3D~\cite{UrbanScene3D} and Mill-19~\cite{MegaNeRF} to increase scene variety.

\paragraph{Urban Data}
Street-view videos, drone footage, and other low-altitude urban imagery are collected and quality-filtered, then matched with the other data sources for joint reconstruction. Public datasets such as UC-GS~\cite{UCGS} provide representative examples, combining drone and ground-level captures within the same urban scenes. Through ABot-3DGS's cross-viewpoint fusion capability (\cref{sec:reconstruction}), these ground-level inputs are registered with aerial data and jointly reconstructed to improve facade detail and novel-view quality at low altitudes.

\begin{table}[t]
\centering
\caption{Open-source real-world datasets used in our data pipeline, covering satellite, aerial, and ground-level viewpoints.}
\label{tab:open_datasets}
\small
\begin{tabular}{lrrlllc}
\toprule
Dataset & Images & Area & DOF & View & Aerial & Depth \\
\midrule
DFC 2019~\cite{DFC2019} & 1K & 25\,km$^2$ & 3 & Satellite & Off-nadir & \checkmark \\
UrbanScene3D~\cite{UrbanScene3D} & 128K & 55\,km$^2$ & 6 & UAV & Any & \checkmark \\
UrbanBIS~\cite{UrbanBIS} & 113K & 10.78\,km$^2$ & 3 & UAV & Any & -- \\
CrossLoc~\cite{CrossLoc} & 57K & 2.7\,km$^2$ & 6 & UAV & Any & \checkmark \\
Mill-19~\cite{MegaNeRF} & 3.5K & 0.18\,km$^2$ & 6 & UAV & Nadir & -- \\
UAVD4L~\cite{UAVD4L} & 0.3K & 2.5\,km$^2$ & 6 & UAV & Any & \checkmark \\
DenseUAV~\cite{DenseUAV} & 27K & -- & 3 & UAV & Nadir & -- \\
UC-GS~\cite{UCGS} & 7K & -- & 6 & UAV+Gnd & Nadir & -- \\
\bottomrule
\end{tabular}
\end{table}

\subsection{City-Scale Reconstruction via ABot-3DGS}
\label{sec:reconstruction}

Converting multi-source imagery into 3DGS representations at this scale poses three main challenges. First, the extreme spatial extent of each project---hundreds of square kilometers---demands scalable computation. Second, heterogeneous scene content (buildings, roads, vegetation, water bodies) requires robust handling across diverse semantics. Third, the multi-source data differ significantly in resolution, viewpoint, and acquisition conditions (time of day, weather, season), introducing substantial appearance variation that must be disentangled from intrinsic scene properties. ABot-3DGS addresses these challenges through the following core capabilities.

\paragraph{Scalable Architecture}
ABot-3DGS employs a hierarchical block-based architecture that partitions city-scale scenes into independently optimizable blocks. A continuous level-of-detail (LOD) hierarchy~\cite{CLOD} adaptively manages scene complexity, and multi-strategy point cloud simplification reduces model size without sacrificing geometric fidelity. The entire pipeline is engineered for GPU cluster parallelism, supporting efficient distributed reconstruction.

\paragraph{Geometry and Detail Optimization}
For geometric accuracy, the pipeline leverages available geometric priors to support reconstruction, and further introduces depth estimation and multi-view geometric consistency to enhance surface precision. For detail enhancement, native training at full input resolution preserves fine-grained textures, and generative refinement recovers appearance in under-observed regions.

\paragraph{Scene Robustness}
Semantics-aware optimization applies differentiated strategies to varied scene content. Multi-level appearance variation modeling disentangles intrinsic scene appearance from transient effects and cross-source inconsistencies such as lighting, weather, and seasonal changes. Dynamic elements such as vehicles and pedestrians are automatically detected and removed.

\paragraph{Cross-View Quality Enhancement}
Cross-view matching enables robust coarse localization and fine registration across data sources captured at vastly different viewpoints. Combined with 3D-consistent image generation, these complementary views are coherently fused into unified reconstructions. Aerial data contributes broad geographic extent while urban captures supply finer-grained details, collectively elevating scene quality beyond what any single source achieves.

Together, these capabilities allow \abotgs to reliably produce photorealistic 3DGS scenes at city scale from large-scale real-world imagery. The resulting reconstructions serve as the data foundation for training our downstream generative model.

\subsection{Training Tile Generation}
\label{sec:training_data}

Given the 3DGS scenes produced by \abotgs, we construct training tile collections through a systematic pipeline. This pipeline converts raw Gaussian primitives into a compact, generation-friendly representation and pairs each tile with dense multi-view supervision.

\paragraph{Spatial Partitioning and Tile Extraction}
A sliding window strategy is applied over the 3DGS scenes, with each tile covering a $200\,\text{m} \times 200\,\text{m}$ region and adjacent tiles overlapping to provide boundary context. Each tile is normalized to a standard coordinate system and cleaned via clustering to remove floating artifacts.

\paragraph{Multi-View Rendering}
Virtual camera arrays are distributed across multiple altitude layers with layer-specific field-of-view settings, covering a range of pitch angles from nadir to oblique viewpoints. Oblique views are sampled at multiple compass directions to capture facade details from all orientations. Random perturbations are applied to camera position, altitude, pitch, and yaw to further increase viewpoint diversity. Additionally, simulated satellite-view images are rendered from the same scenes to provide conditioning inputs for model training (Section~3.4). The resulting rendered views and their source tiles then undergo multi-granularity quality assessment.

\subsection{Data Quality Assessment}
\label{sec:quality}

We establish a multi-granularity quality assessment framework operating at three levels---tile-level reconstruction, view-level rendering, and dataset-level curation---to ensure only reliable, high-quality samples enter the final training set.

\paragraph{Tile-Level 3DGS Reconstruction Assessment}
Each 3DGS tile is evaluated across four dimensions: reference metrics (PSNR / SSIM / LPIPS), geometric accuracy, VLM perceptual quality scores, and spatial completeness. Tiles failing to meet minimum thresholds are recycled to the reconstruction stage with adjusted parameters or excluded entirely.

\paragraph{View-Level Rendering Assessment}
Views with low accumulated opacity are first discarded to eliminate void and boundary regions. A vision-language model (VLM) then scores the remaining views on texture sharpness, artifact absence, and overall perceptual quality. Only views passing both filters are retained as training supervision.

\paragraph{Dataset-Level Curation}
Two curation operations are performed at the tile collection level: spatial diversity balancing, which tracks scene categories and applies stratified sampling so that no single urban morphology dominates; and semantic deduplication, which clusters tiles in embedding space and downsamples near-duplicates to avoid mode collapse.

%% file: sections/sec3_method.tex
\section{Method}
\label{sec3}

While significant progress has been made in object-level 3D generation, with leading methods like TRELLIS~\cite{xiang2025structured}, TRELLIS.2~\cite{xiang2025native}, Hunyuan3D~\cite{lai2025hunyuan3d25highfidelity3d,hunyuan3d22025tencent,yang2024hunyuan3d}, and Seed3D~\cite{Seed3D} demonstrating remarkable capabilities, scaling these paradigms to Earth-scale, real-world outdoor scenes presents a series of fundamental challenges. To address these, our method, \method{}, introduces a tightly integrated set of innovations designed specifically for generative 3D earth modeling, which we detail in the following sections.

\subsection{Native 3DGS Generative Framework}
\label{sec:3dgs_framework}

The first major challenge is a {representation gap}. Existing generators are predominantly designed for clean 3D mesh assets, often created for rendering engines. However, real-world outdoor environments, rich with complex non-manifold topologies like foliage and water surfaces, are more faithfully captured by 3DGS.

To bridge this gap, we pioneer a native 3DGS generative framework. Our approach centers on a compression-generation paradigm that operates directly on the 3DGS representation. It is designed to learn a compact latent space from high-quality, real-world 3DGS scenes, each comprising millions of unstructured Gaussian primitives, and subsequently generate novel scenes directly in this native format. This allows our model to handle the complexity and fidelity of real-world captures without being constrained by mesh-based assumptions.

\subsection{Inherent Multi-LOD Decoding for Interactivity}
\label{sec:lod_decoder}

A second challenge is achieving {scale and interactivity}. Earth-scale generation demands a seamless and continuous Level-of-Detail (LOD) experience, allowing users to transition from a planetary overview to fine-grained, street-level views. This critical interactivity requirement is largely unaddressed by object-centric generators.

Our solution is an inherent multi-LOD decoder that is deeply integrated into the generation process itself. Rather than treating LOD as a post-processing step, our decoder is architected to directly synthesize a hierarchical 3DGS structure. This allows for the on-demand generation of appropriate levels of detail, enabling smooth and real-time online visualization without the overhead of storing or processing multiple discrete versions of the scene.

\subsection{Seamless Sliding-Window Inference for Spatial Coherence}
\label{sec:sliding_window}

Third, ensuring {spatial coherence} at a large scale is paramount. Generating kilometer-scale areas monolithically is computationally prohibitive, but naively stitching multiple generated tiles often results in visible artifacts, breaking the illusion of a continuous world~\cite{blockfusion,ren2024xcube,earthcrafter,Qian_2023_Sat2Density,sat2densitypp,qian2026sat3dgen}.

To overcome this, we propose an efficient seamless sliding-window inference strategy. This mechanism intelligently blends overlapping regions during the generation phase. By carefully managing the influence of adjacent tiles within these transition zones, our strategy drastically reduces stitching artifacts. This makes it practical to render vast, seamless landscapes, ensuring a continuous user experience during large-scale exploration.

\subsection{Cross-Domain Adaptation for Robust Conditioning}
\label{sec:conditioning}

Finally, the model must exhibit {conditional robustness}. Our chosen conditioning signal, satellite imagery, exhibits significant global variance in quality, resolution, and acquisition angles. Furthermore, a substantial domain gap exists between these satellite images and the aerial-view imagery typically used for training reconstructions, primarily due to atmospheric effects and different sensor characteristics.

To ensure our model performs reliably, we employ a cross-domain conditional adaptation strategy. This is a two-stage approach. During training, we simulate satellite-view renderings from our training data to provide a consistent conditional input for the model to learn from. At inference, we introduce a novel Vision-Language Model (VLM)-based harness that dynamically adapts the conditioning to the specific characteristics of the real-world satellite input. This ensures robust and high-fidelity 3D content generation from any real-world satellite image, regardless of its origin or quality.

In summary, through this tightly integrated set of innovations, \method{} achieves the first end-to-end generation of interactive, streamable, Earth-scale 3D outdoor environments directly from real-world satellite imagery.

%% file: sections/sec4_deployment.tex
\section{Deployment: From Algorithm to a Planetary-Scale System}
\label{sec:deployment}

Bridging the gap between a novel generative algorithm and a robust, planetary-scale service requires a sophisticated engineering deployment strategy. The theoretical power of \method{} is realized through a two-stage, end-to-end pipeline designed for massive scalability and real-time performance. The first stage is a {Global-Scale Production Pipeline} responsible for generating trillions of Gaussian primitives from satellite imagery. The second stage is a {Scalable Post-Processing and Rendering Pipeline} that organizes this colossal dataset for interactive, real-time exploration on a map engine. This section details the architecture and key engineering decisions of this entire system.

% ---------------------------------
\subsection{Global-Scale 3DGS Production Pipeline}

\paragraph{Tile-Based Generation Strategy and Resource Planning}
To operationalize global-scale scene generation, we designed a large-scale, tile-based concurrent production pipeline. The primary engineering constraint is the VRAM capacity of inference GPUs, which dictates the maximum input image resolution and, consequently, the spatial extent of a single generation run. Our strategy partitions the global target area into regular spatial tiles, with each tile processed as an independent generation task.

Using A100 GPUs, a single inference pass can process a 4K resolution satellite image, corresponding to a ground coverage of approximately 1.6km × 1.6km (2.56km²). This inference scale represents a {64-fold increase in area} compared to our 200m × 200m training tiles, demanding exceptional long-range spatial consistency from our model. Our current strategy focuses on achieving near-perfect seamlessness within each 1.6km × 1.6km block, leveraging an internal sliding-window mechanism. This modular, block-based approach is a pragmatic choice to maximize internal quality under current hardware limitations. Based on a global built-up area of approximately 800,000km², this results in a total of roughly 312,500 production tiles. A fully seamless, cross-block inference strategy is theoretically feasible and is a key objective for future iterations of our pipeline.
% ---------------------------------

\paragraph{Production Scheduling and Performance}
Under a 1,000-GPU cluster configuration, single-tile inference completes in approximately 25 minutes. The entire production run, comprising over 300 concurrent batches, is estimated to finish in under 10 days. Our task scheduling system is designed for robustness, employing dynamic queues and load-balancing to mitigate the impact of straggler tasks caused by varying scene complexity (e.g., dense urban cores vs. sparse suburbs). The pipeline supports checkpoint-based resumption and automatic retries, ensuring the reliability of this massive and long-running production job. The final output, comprising hundreds of billions of Gaussian primitives, presents a substantial yet manageable data organization challenge.
% ---------------------------------

\paragraph{Georeferencing and Input Preprocessing}
Accurate geospatial alignment is critical. We record the geographic bounding box for each tile, providing the necessary anchors for global coordinate transformation. A critical preprocessing step addresses the scale variance of input imagery. Since our model is trained within a specific Ground Sampling Distance (GSD) range, we uniformly rescale all input images to match this training scale, ensuring stable geometric and textural synthesis.

This is particularly important when handling standard Web Mercator (EPSG:3857) raster tiles, which suffer from significant areal distortion at higher latitudes. To counteract this distortion and ensure a consistent input scale, our pipeline first mosaics the Web Mercator tiles into a contiguous geographic image. It then performs an isotropic resampling process based on the target spatial extent, guaranteeing a uniform effective GSD across all latitudes. This rigorous preprocessing satisfies the model's dimensional and scale constraints, ensuring high-quality generation globally.
% ---------------------------------
% \begin{figure}[htp]
%     \centering
%     \includegraphics[width=0.5\linewidth]{sections/CutBlock.png}
%     \caption{The tile-based production pipeline partitions the globe for parallel inference.}
%     \label{fig:cutblock}
% \end{figure}

% \begin{figure}[htp]
%     \centering
%     \includegraphics[width=0.5\linewidth]{sections/LODTiles.png}
%     \caption{Hierarchical LOD tiles are generated for efficient, multi-resolution streaming.}
%     \label{fig:lodtiles}
% \end{figure}

% \begin{figure}[htp]
%     \centering % Center the entire figure block

%     % --- First Subfigure ---
%     % Note: We now specify height instead of width
%     \begin{subfigure}{\dimexpr0.48\linewidth} % The width is just a container, can be adjusted
%         \centering
%         \includegraphics[height=4cm]{sections/CutBlock.png} % Specify a fixed height
%         \caption{The tile-based production pipeline partitions the globe for parallel inference.}
%         \label{fig:cutblock}
%     \end{subfigure}
%     \hfill % Adds flexible horizontal space
%     % --- Second Subfigure ---
%     \begin{subfigure}{\dimexpr0.48\linewidth}
%         \centering
%         \includegraphics[height=4cm]{sections/LODTiles.png} % Use the EXACT SAME height
%         \caption{Hierarchical LOD tiles are generated for efficient, multi-resolution streaming.}
%         \label{fig:lodtiles}
%     \end{subfigure}

%     \caption{Overview of the production and data organization pipeline.}
%     \label{fig:pipeline_overview}
% \end{figure}

\begin{figure}[ht]
    \centering % Center the entire figure block

    % --- First Subfigure ---
    \begin{subfigure}[b]{0.45\linewidth}
        \centering
        % 直接为图片指定一个固定的高度，例如 4cm
        \includegraphics[height=6cm]{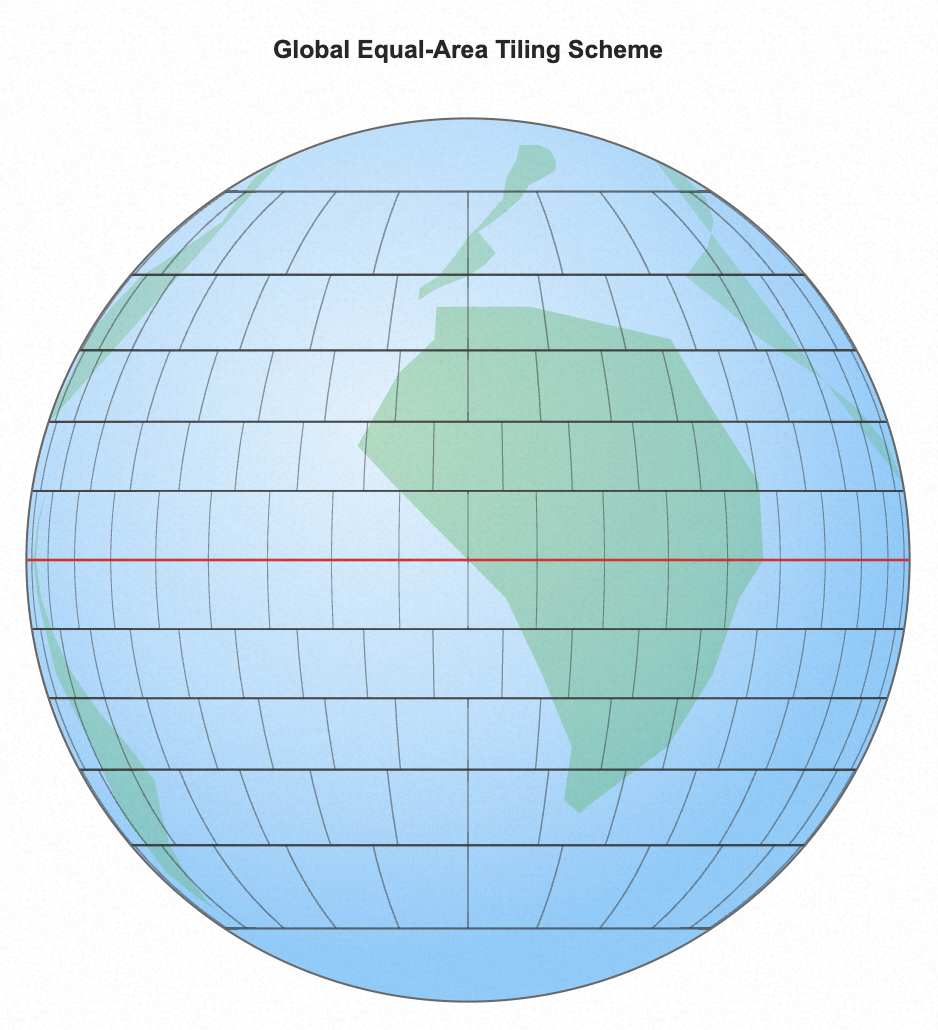} 
        \caption{The tile-based production pipeline partitions the globe for parallel inference.}
        \label{fig:cutblock}
    \end{subfigure}
    \hfill % 在两图之间添加弹性空白
    % --- Second Subfigure ---
    \begin{subfigure}[b]{0.51\linewidth}
        \centering
        % 使用与第一张图完全相同的高度值
        \includegraphics[height=6cm]{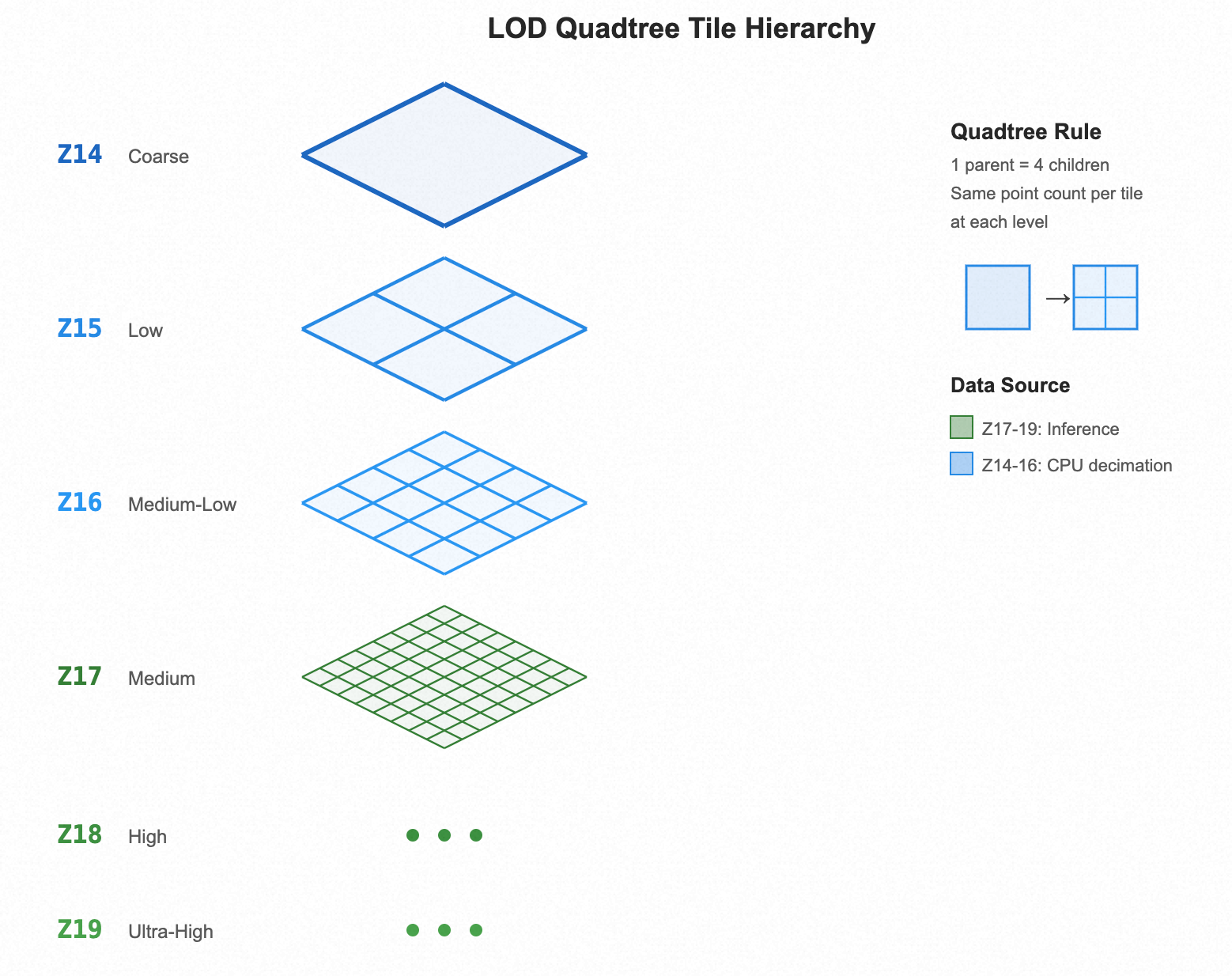}
        \caption{Hierarchical LOD tiles are generated for efficient, multi-resolution streaming.}
        \label{fig:lodtiles}
    \end{subfigure}

    \caption{Overview of the production and data organization pipeline.}
    \label{fig:pipeline_overview}
\end{figure}

\subsection{EarthScape: Scalable Rendering Pipeline}

The production pipeline yields approximately 320,000 inference blocks, which contains approximately 3.2 trillion Gaussian primitives. This colossal data volume presents two fundamental engineering bottlenecks: (1) the 100 million primitives per block far exceed the rendering capacity of consumer GPUs, and (2) each block exists in an independent local coordinate system, preventing direct assembly into a continuous scene. Our deployment pipeline addresses these challenges through three core pillars: geographic alignment, LOD data reorganization, and rendering scheduling.
% ---------------------------------

\paragraph{I. Geographic Alignment: Unified Coordinate Transformation}
Each inference block is normalized into a unified coordinate framework. First, using affine transformation parameters saved during production, each model is restored to its projected coordinate space (EPSG:3857). Subsequently, an ENU (East-North-Up) local tangent plane coordinate system is established at the tile's center. All Gaussian primitives—including their positions, rotation quaternions, and scaling parameters—are uniformly transformed into this ENU frame. This process ensures that all block models share a common, meter-scale coordinate system suitable for rendering engines, while retaining precise global georeferencing through their ENU origins. This alignment establishes the spatial datum for all subsequent processing.

\paragraph{II. LOD Data Reorganization: A Prerequisite for Interaction}
Given the 3.2 trillion primitive dataset, Level-of-Detail (LOD) organization is not an optimization but an absolute prerequisite for deployment. Our solution is a multi-pronged strategy:

\textit{Tile Re-partitioning.} After alignment, all Gaussians are re-assigned to a standard map tile hierarchy (zoom/x/y), merging data across inference block boundaries. This creates a 6-level LOD structure spanning from zoom level 14 to 19.

\textit{Multi-level LOD Generation.} The three highest-precision levels (zoom 17-19) are generated natively by the inference model itself, which supports multi-resolution outputs. This avoids quality loss from downsampling. The three lower-precision levels (zoom 14-16) are generated from the zoom-17 data using a statistical decimation scheme guided by the Bhattacharyya distance. This method is highly efficient as it operates analytically on Gaussian parameters, allowing us to offload this task to CPUs and run it in parallel with GPU-based inference, significantly reducing end-to-end latency by leveraging heterogeneous compute resources.

\textit{Spatial Indexing.} We construct a two-tier spatial index: an explicit `tileset.json` compliant with the Open Geospatial Consortium (OGC) 3D Tiles specification \cite{ogc2023referencing} for standard clients, and an implicit map tile path convention (\texttt{\{zoom\}/\{x\}/\{y\}}) for direct access and Content Delivery Network (CDN) caching.

\paragraph{III. Rendering Scheduling: Real-time Rendering via Map Engine}
This entire data organization strategy culminates in its integration with the Amap Yunjing rendering engine. The engine's native tile scheduling capabilities are leveraged to manage the 3DGS data. Each frame, it dynamically computes the required tile set and precision level based on the camera's viewport. Close views load high-precision tiles (zoom 17-19), while distant views load coarse-grained tiles (zoom 14-15), with smooth fade transitions between levels. This approach reuses the engine's existing frustum culling and asynchronous streaming infrastructure, ultimately achieving real-time, interactive rendering of a trillion-scale global 3DGS dataset.

\begin{figure}[htp]
    \centering
    \includegraphics[width=1\linewidth]{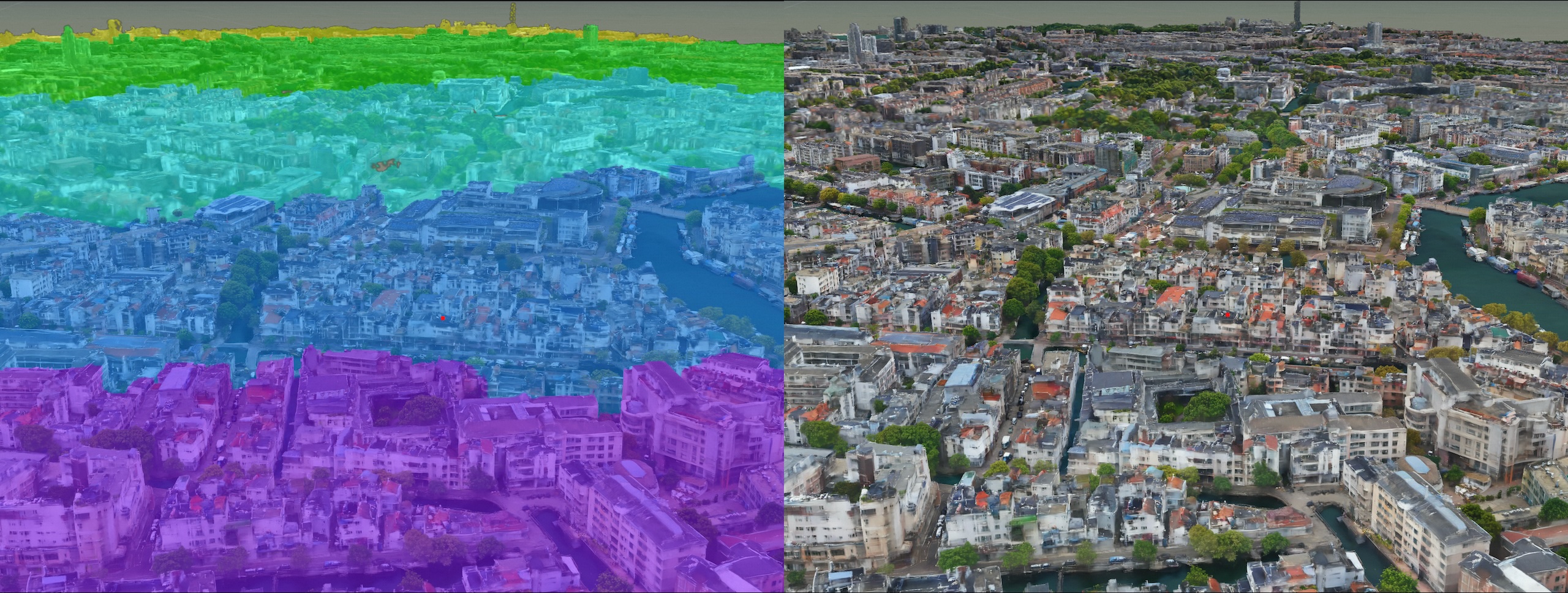}
    \caption{LOD rendering in the map engine, enabling seamless exploration from global to street-level views.}
    \label{fig:rendered}
\end{figure}

%% file: sections/sec5_evaluation.tex
\section{Evaluation}
\label{sec:result}

We evaluate our method from two complementary perspectives: generative fidelity and system-level applicability.

To assess generative fidelity, we conduct a quantitative comparison against existing academic baselines for outdoor scene generation, thereby establishing the fundamental realism of our model. To evaluate system-level applicability, we examine our framework from multiple angles. We first perform a product-level benchmark against leading commercial solutions, focusing on performance and efficiency. Furthermore, we demonstrate the practical utility and extensibility of our system by showcasing a hybrid generation-reconstruction approach for integrating high-fidelity landmark models, a key feature for real-world applications.

\subsection{Generative Fidelity}

As presented in \cref{tab:conditions_results}, we assess the generative fidelity of our method against prominent outdoor scene generation baselines: CityDreamer~\cite{CityDreamer}, GaussianCity~\cite{Xie_2025_CVPR}, and EarthCrafter~\cite{earthcrafter}. The evaluation employs the standard FID~\cite{NIPS2017_8a1d6947} and KID~\cite{binkowski2021demystifyingmmdgans} metrics, which are computed between a set of our generated 2D renderings and a ground-truth image set derived from real-world data. Baseline results are cited from their respective publications.

Our method achieves a state-of-the-art FID score of 16.1, a substantial improvement over the previous best of 69.5. This result is particularly compelling because our metrics are benchmarked against a ground-truth distribution derived from renderings of highly complex, real-world 3DGS reconstructions. This poses a significantly greater modeling challenge compared to evaluations on more constrained or synthetic datasets, underscoring our model's superior capability to capture the photorealism and intricate detail of authentic aerial environments.

Crucially, our contribution extends far beyond superior rendering quality. While existing research primarily focuses on generating isolated or limited-area scenes, our work is the first to directly tackle the challenges of creating continuous, interactive, Earth-scale 3D environments. \method{} therefore not only sets a new benchmark for generative fidelity but also marks a fundamental step towards planetary-scale digital twins, a frontier previously unexplored by these methods.

\input{body/table_tex/tab_fid}

\subsection{System-level Applicability}
\label{sec:system_applicability}

While academic benchmarks focus primarily on pixel-level generative fidelity, deploying 3D generation models in real-world pipelines requires a holistic evaluation of system-level performance. To this end, we conduct a comparative analysis between \method{} and two leading commercial solutions: Google Earth (the industry standard for photogrammetry-based planetary reconstruction) and Marble (a state-of-the-art closed-source procedural 3D world-generation platform).

We evaluate these systems across four key dimensions critical for large-scale production: \textit{Spatial Coverage}, \textit{Timeliness and Efficiency}, \textit{Visual Quality}, and \textit{System Openness}. A high-level overview of this multi-dimensional comparison is summarized in \cref{tab:system_comparison_merged}, with detailed analyses in the following sections.

\input{body/table_tex/earth_marble_compare}

\subsubsection{Spatial Coverage and Scalability}
A system's value is fundamentally tied to its reach. Here, \method{}'s generative paradigm offers a significant advantage. As shown in the continental-scale comparison in \cref{fig:system_comparison}, Google Earth's coverage is constrained by physical acquisition; its high-fidelity 3D geometry is limited to well-surveyed metropolitan areas. According to its official platform~\footnote{\url{https://developers.google.com/maps/coverage\#countryregion-coverage}}, 3D assets cover only a fraction of global countries, and even within these regions, data is sparse, often omitting non-CBD areas. In contrast, by leveraging a continuous latent space, \method{} rapidly provides 3D assets for vast regions, including numerous developing countries and smaller cities. As illustrated in \cref{fig:coverage}, \method{} successfully generates a plausible 3D scene of Ireland, whereas Google Earth, lacking scan data for this region, falls back to a 2D image.

\input{figure/fig_tex/compaer_googleearth}

\subsubsection{Efficiency}
In terms of efficiency, \method{} demonstrates clear superiority. Requiring only satellite imagery as input, it can generate 1 km² in under 10 minutes, offering exceptional timeliness for on-demand 3D environment creation. In stark contrast, commercial photogrammetry pipelines are slow, batch-processed endeavors. Changes in Google Earth's geometry typically require months to years to propagate from image capture to renderer.  This makes \method{} uniquely suited for applications requiring rapid updates or synthesis of unmapped areas.

\input{body/table_tex/coverage_and_score}

\subsubsection{Visual Quality and Aesthetics}
Beyond performance metrics, we also conducted a comprehensive human study to assess visual quality, with results summarized in the radar chart in \cref{fig:system_comparison}. We defined three components for participants: Geometric Accuracy (structural integrity), Textural Fidelity (surface detail representation), and Overall Aesthetics (holistic appeal, including lighting and color harmony).

Notably, \method{} achieves a higher aesthetic score than Google Earth. We attribute this to the holistic nature of aesthetic judgment: observers often prioritize plausible lighting and color harmony over micro-level texture accuracy. Our model excels at replicating these holistic photorealistic qualities.

Conversely, and as anticipated, Google Earth maintains advantages in geometric and textural fidelity. This is an expected outcome, given that Google's reconstruction algorithms have been meticulously optimized over many years, incorporating sophisticated strategies like optimized aerial survey patterns, strong priors ("Manhattan-world" assumption), and extensive manual post-processing. We liken this current gap to the difference between a 3D model hand-crafted by a professional artist and one from a first-generation generative model (e.g., LRM~\cite{hong2024lrm} or CLAY~\cite{CLAY}).

However, we believe this quality gap is not fundamental. With the continued evolution of \method{}, we are confident that our generative capabilities will progressively close this gap, with the potential to eventually approach and even exceed the quality of traditionally reconstructed outdoor scenes.

\subsubsection{System Openness and Downstream Integration}
Commercial solutions often operate as closed-loop proprietary ecosystems. Google Earth restricts users to its viewer or constrained API tiles, prohibiting direct access to raw data. Marble operates as a black-box service with limited export pipelines. Conversely, \method{} is built on open standards. Our model generates native 3DGS that can be rendered from any angle, facilitating immediate downstream utility in simulation, virtual production, and spatial computing.

\begin{figure}[ht]
    \centering
    \includegraphics[width=0.95\linewidth]{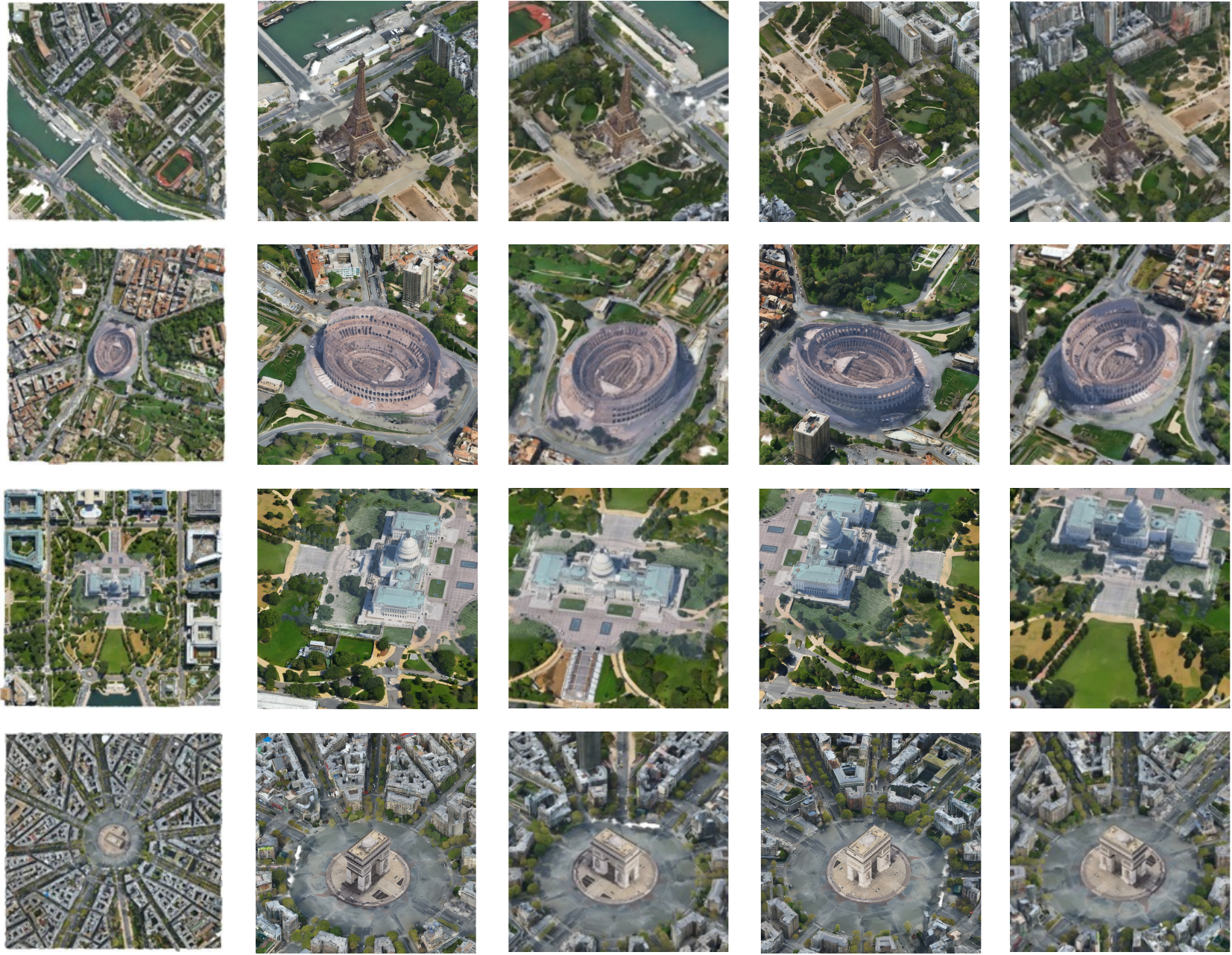}
    \caption{Landmark integration results. We composite reconstructed landmarks into our generative environments. Top-down views are in the leftmost column, with oblique renderings in the others. From top to bottom: Eiffel Tower, Colosseum, US Capitol, and Arc de Triomphe. The models preserve fine-grained architectural details and blend effectively with the surrounding context.}
    \label{fig:dibiao}
\end{figure}

\subsection{Landmark Enhancement: Exploring a Hybrid Generative-Reconstructive Approach}

An interesting distinction arises when considering the generation of iconic landmarks. While a typical generated city block might be evaluated on its overall realism, a world-renowned structure like the Eiffel Tower or the Roman Colosseum is often compared directly to a user's well-established mental image. This suggests that enhancing such landmarks with higher-fidelity models could significantly improve the overall user experience and sense of place. Simply scaling up generative efforts for this task is often impractical, motivating an exploration into a hybrid ``best-of-both-worlds'' paradigm that is both economical and targeted. This approach leverages fast, scalable generation for the vast surrounding context while integrating dedicated, high-fidelity reconstructions only where they might offer the most impact.

To investigate this possibility, we experimented with a classical Structure-from-Motion and Multi-View Stereo pipeline (COLMAP)~\cite{schoenberger2016sfm,schoenberger2016mvs} on crowd-sourced imagery to create dense reconstructions of selected landmarks. These were then converted into our native 3D Gaussian Splatting format, geo-registered, and composited into the generatively synthesized environment. The results of this integration are shown in \cref{fig:dibiao}, which includes several high-fidelity landmark models.

This exploration into hybrid scenes hints at a core principle of ABot-Earth: its potential as an editable and extensible platform. A scene generated by ABot-Earth need not be a static, immutable image but can serve as a structured spatial foundation for secondary creation and information fusion. For urban planners, this could mean clearing entire blocks to visualize future design blueprints. For emergency commanders, it opens possibilities for overlaying dynamic data like fire progression onto a 3D command sandbox. For business clients, it could enable the fusion of logistics routes and demographic heatmaps for operational analysis. This potential to merge a generative base with dynamic data streams points toward a future where ABot-Earth evolves from a map into a spatial intelligence and decision-making platform for diverse industries.

\input{body/table_tex/large_scale_metric}
\input{body/table_tex/large_scale_quality}

%% file: body/table_tex/tab_fid.tex
\begin{table}[h]
\caption{Quantitative comparison on images. FID/KID values for baselines are computed using different GT sets than ours. In addition, the poses/viewpoints used for evaluation differ across methods (e.g., different near/far sampling). The reported metrics are for reference only.}
\centering
\small
\setlength{\tabcolsep}{10pt}
\begin{tabular}{lcc}
\toprule
\textbf{Method} & \textbf{FID} & \textbf{KID} \\
\midrule
CityDreamer~\cite{CityDreamer} & 97.3 & 0.096 \\
GaussianCity~\cite{Xie_2025_CVPR} & 86.9 & 0.090 \\
EarthCrafter~\cite{earthcrafter} & 69.5 & 0.061 \\
\hline
Ours &\textbf{16.1}&\textbf{0.006}\\
\bottomrule
\end{tabular}
\label{tab:conditions_results}
\end{table}

%% file: body/table_tex/earth_marble_compare.tex
\begin{table}[t]
\centering
\caption{\textbf{System-level and technical comparison with commercial baselines.} We compare \method{} against leading industrial solutions across key technical and system-level dimensions.}
\label{tab:system_comparison_merged}
\small
\begin{tabular}{l ccc}
\toprule
\textbf{Dimension} & \textbf{Google Earth} & \textbf{Marble} & \textbf{\method{} (Ours)} \\
\midrule
\textbf{Paradigm} & Reconstruction & Generation & {Generation} \\
\textbf{Coverage} & Sparse (Scanned region only) & N/A & {Infinite} \\
% \textbf{Timeliness} & Years & N/A & {Hours} \\
% \textbf{Georeferenced} & \checkmark (Real-world coordinates) & \times & \textbf{\checkmark (Prompt/Layout-guided)} \\
% \addlinespace
% \textbf{Resource Cost} & Extremely High & Low & \textbf{Low}  \\
\textbf{Openness} & API only & Open Platform & {Open Platform} \\
\bottomrule
\end{tabular}
\end{table}

%% file: figure/fig_tex/compaer_googleearth.tex
% 引用：https://developers.google.com/maps/coverage?hl=zh-cn#countryregion-coverage

\begin{figure}[htp]
\centering
% === Group 1: Auckland (Comparable Quality) ===
\begin{subfigure}[t]{0.38\linewidth}
\centering
\includegraphics[width=\linewidth]{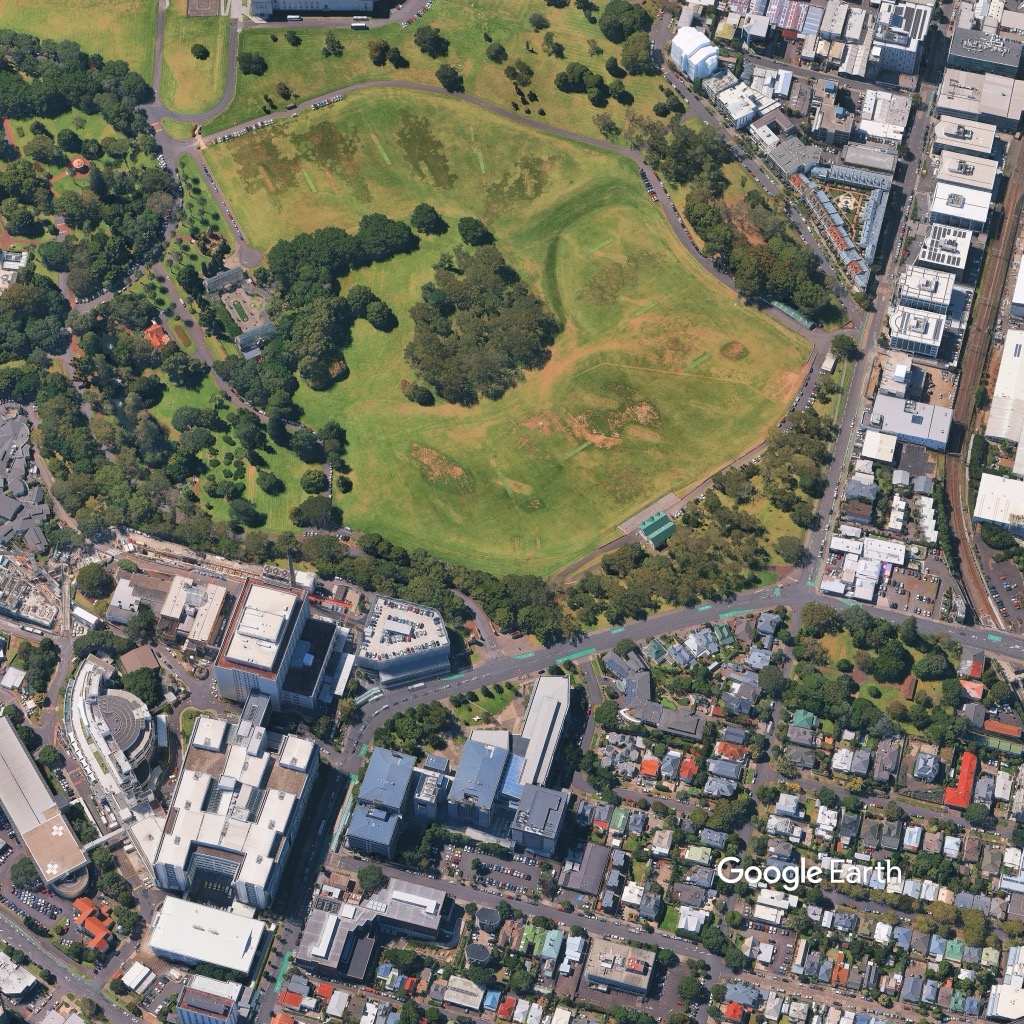}
\caption{Google Earth (New Zealand)}
\label{fig:auckland_google}
\end{subfigure}
\hspace{15pt} % <--- 将 \hfill 改为固定间距，15pt 可以根据需要微调（数值越小越靠拢）
\begin{subfigure}[t]{0.38\linewidth}
\centering
\includegraphics[width=\linewidth]{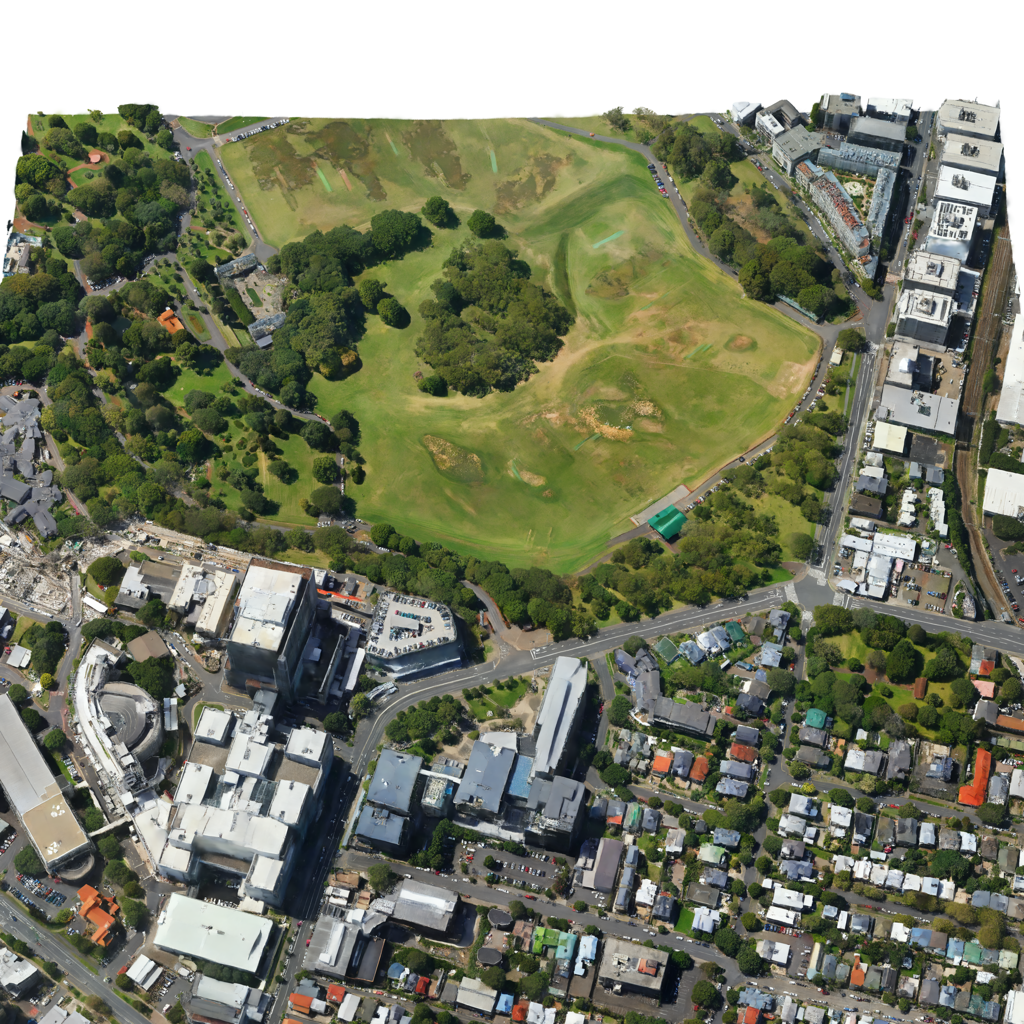}
\caption{\method{} (New Zealand)}
\label{fig:auckland_ours}
\end{subfigure}

\vspace{6pt}

% === Group 2: US Capitol (Slightly Weaker) ===
\begin{subfigure}[t]{0.38\linewidth}
\centering
\includegraphics[width=\linewidth]{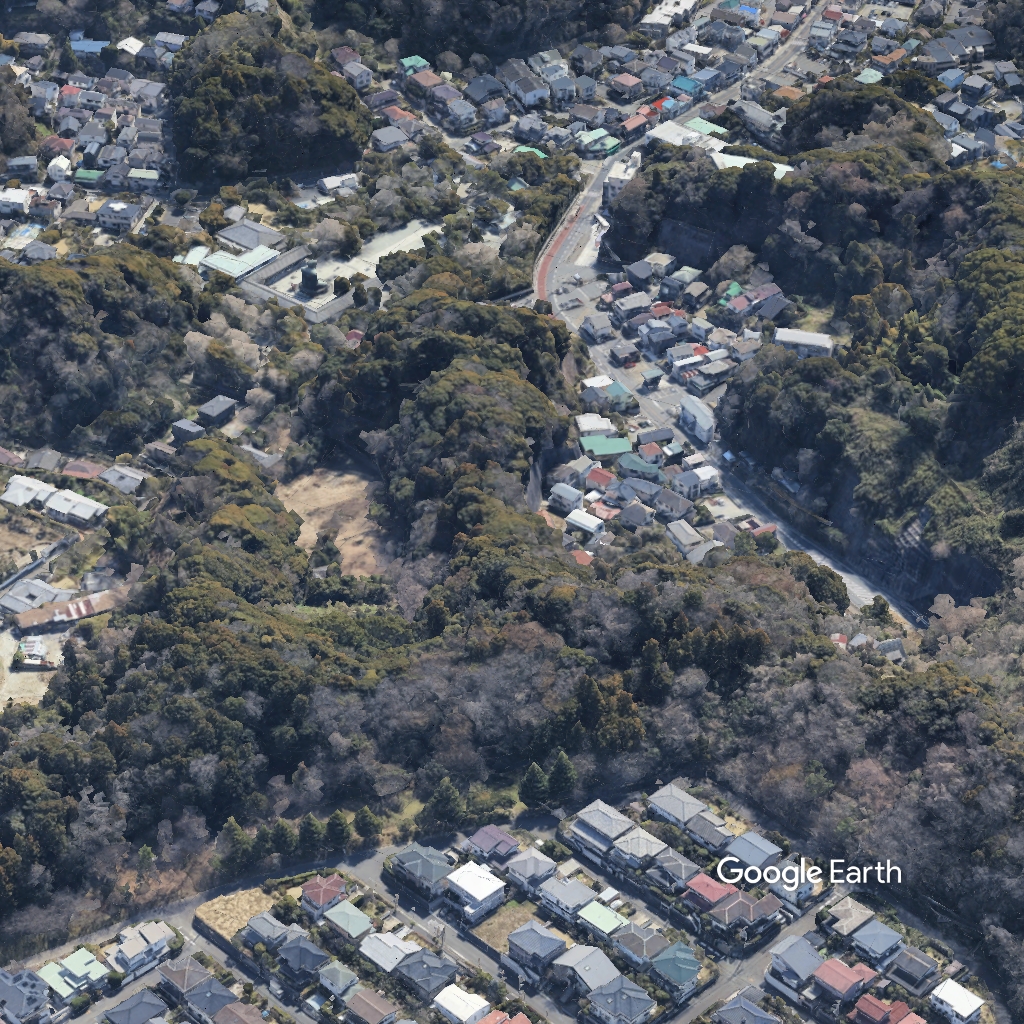}
\caption{Google Earth (Japan)}
\label{fig:capitol_google}
\end{subfigure}
\hspace{15pt} % <--- 同步修改
\begin{subfigure}[t]{0.38\linewidth}
\centering
\includegraphics[width=\linewidth]{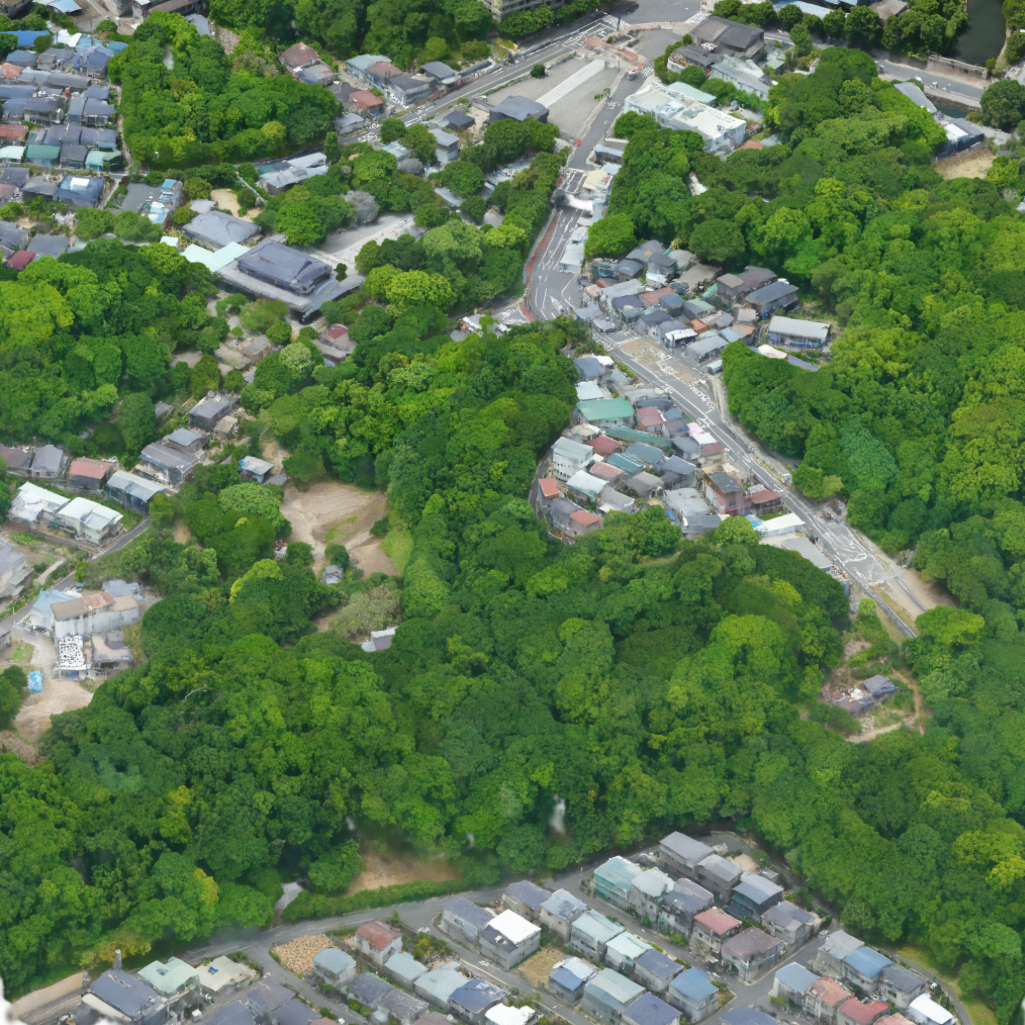}
\caption{\method{} (Japan)}
\label{fig:capitol_ours}
\end{subfigure}

\vspace{6pt}

% === Group 3: Istanbul (No Google 3D vs. Ours) ===
\begin{subfigure}[t]{0.38\linewidth}
\centering
\includegraphics[width=\linewidth]{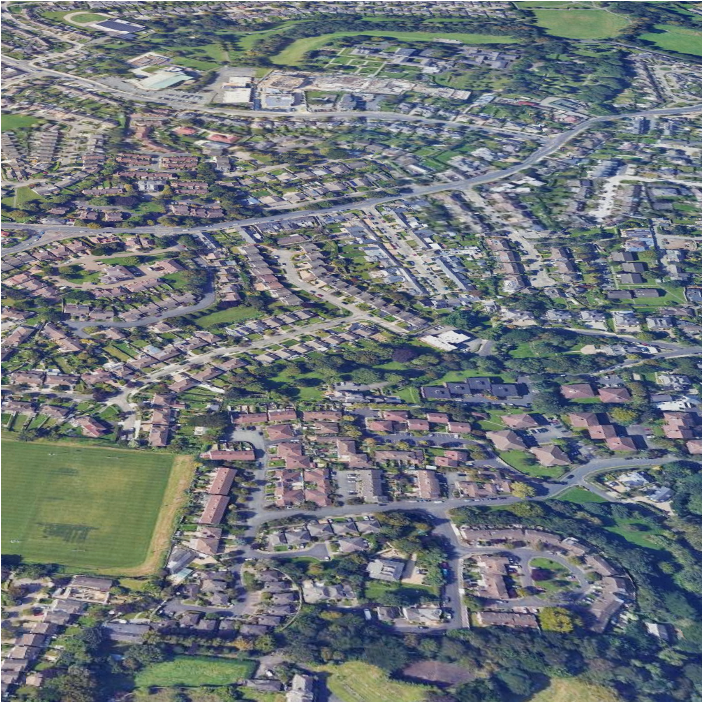}
\caption{Google Earth (Ireland,  no 3D tile)}
\label{fig:istanbul_google}
\end{subfigure}
\hspace{15pt} % <--- 同步修改
\begin{subfigure}[t]{0.38\linewidth}
\centering
\includegraphics[width=\linewidth]{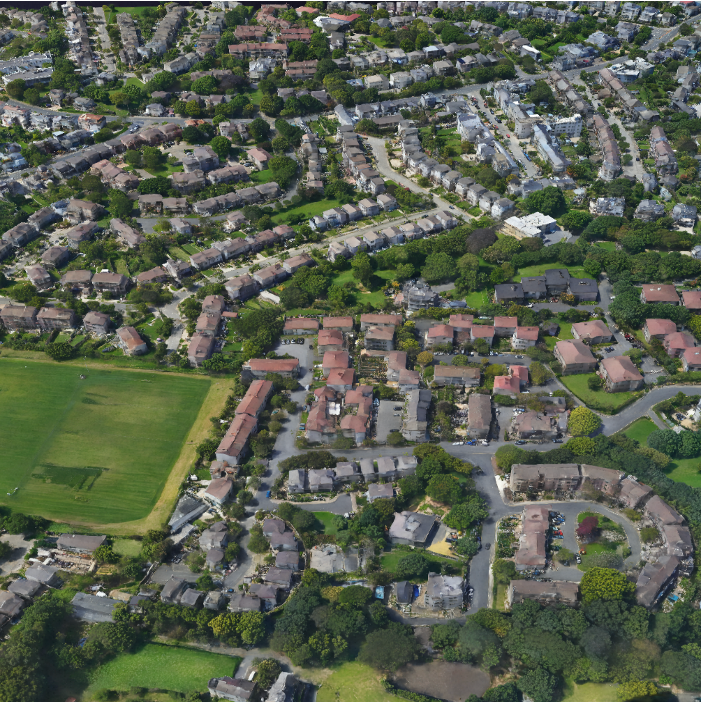}
\caption{\method{} (Ireland)}
\label{fig:istanbul_ours}
\end{subfigure}

\caption{Qualitative comparisons between Google Earth and \method{} across different regions: 
{(a-b) New Zealand} and {(c-d) Japan}, where our method achieves comparable visual quality to Google Earth's 3D reconstruction; and 
{(e-f) Ireland}, where Google Earth lacks 3D scanning data and renders only a flat map approximation, while \method{} successfully synthesizes a detailed 3D scene from satellite imagery.}

\label{fig:coverage}
\end{figure}

%% file: body/table_tex/coverage_and_score.tex
% ===================================================================
%  并排的子图（(a) 总体比较/雷达图，(b) 大洲覆盖率）
% ===================================================================
\begin{figure}[t!]
    \centering % 将 figure 内的所有内容整体居中

    % ----- (a) 总体比较：雷达图 -----
    \begin{subfigure}[c]{0.48\linewidth}
        \centering
        \includegraphics[width=1.0\linewidth]{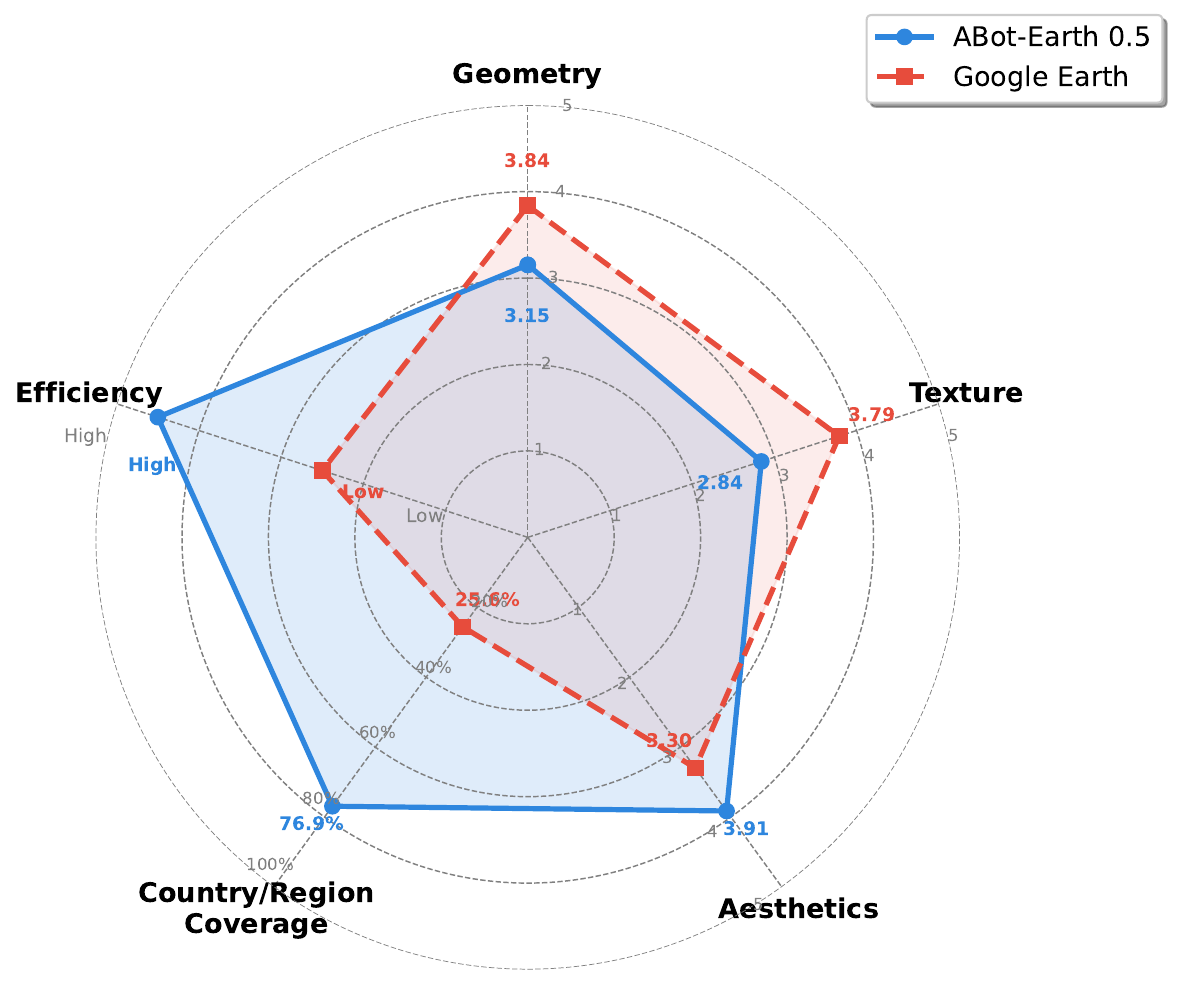}
        \caption{Overall comparison.} % 子图 (a) 的小标题
        \label{fig:radar_metrics}           % 子图 (a) 的引用标签
    \end{subfigure}
    \hfill % 在左右两部分之间添加一个弹性的空白
    % ----- (b) 大洲覆盖率图表 -----
    \begin{subfigure}[c]{0.51\linewidth}
        \centering
        \includegraphics[width=1.0\linewidth]{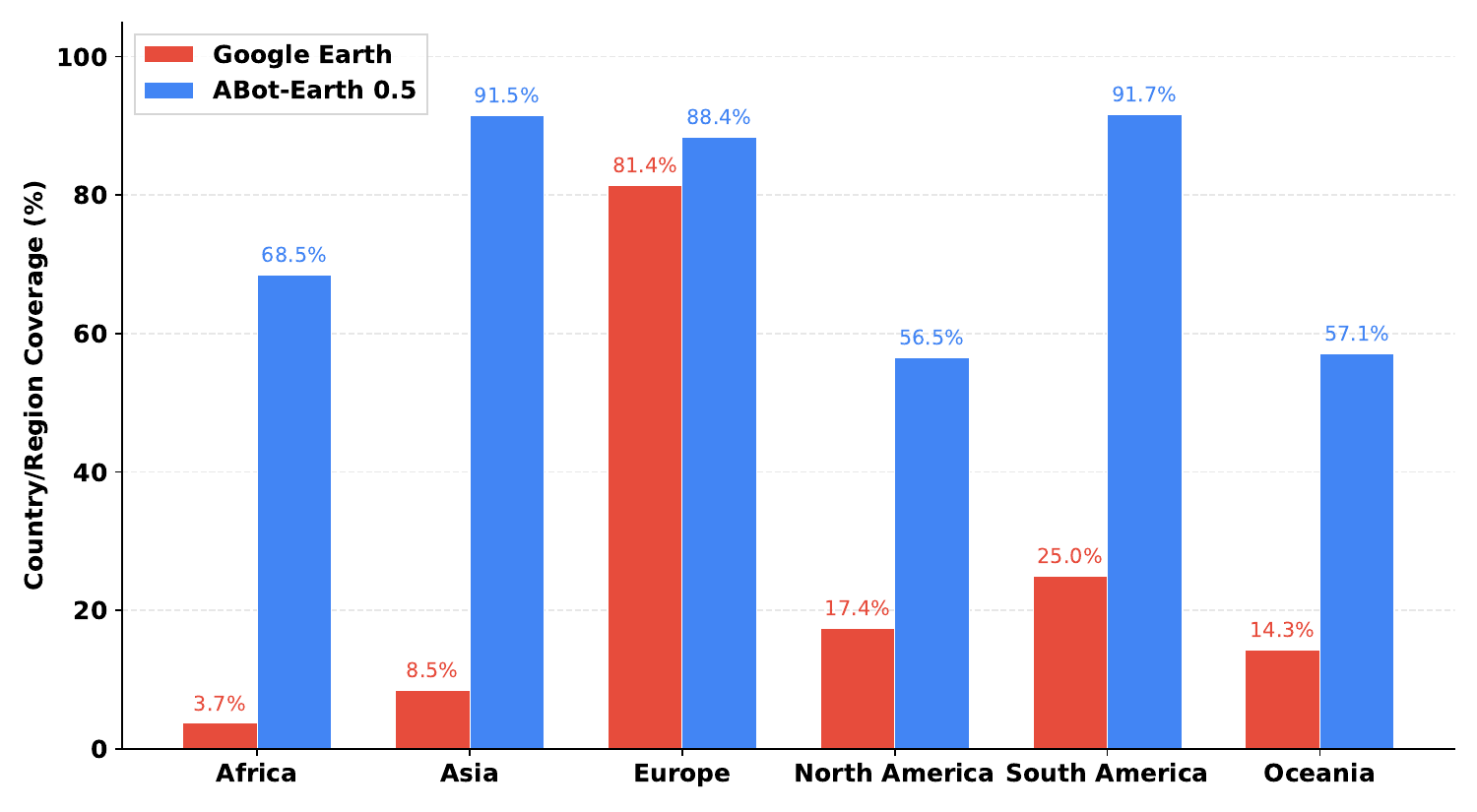}
        \caption{Continental 3D coverage comparison.} % 子图 (b) 的小标题
        \label{fig:continent_coverage}      % 子图 (b) 的引用标签
    \end{subfigure}

    % 更新后的总标题
    \caption{Comparative analysis of \method{} and Google Earth. (a) Overall comparison of key system metrics and user-rated visual quality, including geometry, texture, and aesthetics. (b) Continental 3D coverage comparison, highlighting the broader reach of \method{} across all continents.}
    \label{fig:system_comparison} % 给整个 figure 一个总引用标签

\end{figure}
% ===================================================================

%% file: sections/sec6_conclusion.tex
\section{Conclusion}
\label{sec:conclusion}

The core value of ABot-Earth is to democratize 3D content production, transforming it from a high-cost, specialized process into a low-barrier, generative workflow. It rapidly generates foundational 3D layers for urban digital twins, significantly accelerating project timelines, and fills the void in unmapped regions globally, enabling the low-cost launch of 3D map services. Furthermore, ABot-Earth provides a crucial spatial prior for intelligent systems such as drones, enabling advanced navigation and simulation in complex environments. It also turns physical locations into versatile creative assets, allowing for the easy generation of a scenic area's seasonal variations, a city's future developments, or narrative-driven spaces for individual creators. These applications all converge on a single trend: the future of the digital Earth will transcend being a passive viewing tool to become a dynamic spatial intelligence platform—one that is generative, simulatable, and editable. ABot-Earth marks the beginning of this evolution.

Looking ahead, our next step is to bring this technology from the sky down to the ground~\cite{qian2026sat3dgen}. We are working to transition from our current aerial-level 3D to street-view level detail, while simultaneously exploring greater scene diversity and achieving reconstruction-grade fidelity in our outputs. Concurrently, we aim to systematically validate the scaling laws that govern outdoor 3D scene generation. Our vision is for ABot-Earth to become the foundational layer for the future 3D world, ensuring that a new generation of applications, from digital twins to robotics simulation, can all benefit and build upon it.

%% file: sections/contribution.tex
\section{Contributions}
%  this style follows lingbot-world

\textbf{Algorithm:} Ming Qian,  Tianjian Ouyang,  Mingchao Sun,  Zijian Wang, Jincheng Xiong, Jiarong Han, Yongchang Zhang,  Jiawei Zhang

\textbf{Data Pipeline:}  Mingchao Sun, Yongchang Zhang,  Zijian Wang, Xu Wang, Yu Liu, Luyang Tang, Zengye Ge

\textbf{Engineering:} Mengmeng Du, Yuan Liu,  Nianfei Fan, Song Wang, Yingliang Peng

\textbf{Art Designer:} Chunxue Jia, Yang Liu, Shiying Zeng,  Haozhe Shi

\textbf{Project Sponsor:} Mu Xu,  Junnan Lai,  Hongyu Pan,  Zheng Wu, Ning Guo
% , Ning Guo

\textbf{Project Leader:} Hang Zhang, Ming Qian, Mingchao Sun

We would like to express our sincere gratitude to Jian Zhang, Yu Lei, Chong Sun, and Qianwei Wang for their valuable support and contributions to this project.
% no * or multiple *  are also acceptable.

% \begin{multicols}{-2}
% \subsubsection*{Core Contributors}
%     \begin{itemize}

%         \item xxx$^\dagger$

%     \end{itemize}

%     \columnbreak % Force the remaining content to the second column
% % \cleanpage
%     \subsubsection*{Contributors}
%     \begin{itemize}
%         % contributor
%         \item xxx
%     \end{itemize}
% \end{multicols}

% {\renewcommand{\thefootnote}{\fnsymbol{footnote}}\footnotetext[2]{Corresponding author: xumu.xm@alibaba-inc.com}}